\documentclass[square,preprint,review,12pt]{elsarticle}

\usepackage{amssymb}
\usepackage{amsmath}

\usepackage{booktabs}
\usepackage{soul}
\usepackage{makecell}
\usepackage{bbding}
\usepackage{pifont}
\usepackage{bigstrut}
\usepackage{color}
\usepackage{algorithm}
\usepackage{algpseudocode}
\usepackage{multirow}
\usepackage[colorlinks,linkcolor=blue]{hyperref}

\journal{Nuclear Physics B}

\begin{document}

\begin{frontmatter}



\title{BALR-SAM: Boundary-Aware Low-Rank Adaptation of SAM for Resource-Efficient Medical Image Segmentation}

\author[label1]{Zelin Liu}
\author[label1]{Sicheng Dong}
\author[label1]{Bocheng Li}
\author[label1]{Yixuan Yang}
\author[label1]{Jiacheng Ruan}
\author[label1]{Chenxu Zhou}
\author[label1]{Suncheng Xiang}

\affiliation[label1]{organization={Shanghai Jiao Tong University},
            addressline={No 800, Dongchuan Road, Minhang District},
            city={Shanghai},
            postcode={200240},
            state={Shanghai},
            country={China}}



\begin{abstract}
Vision foundation models like the Segment Anything Model (SAM), pretrained on large-scale natural image datasets, often struggle in medical image segmentation due to a lack of domain-specific adaptation. In clinical practice, fine-tuning such models efficiently for medical downstream tasks with minimal resource demands, while maintaining strong performance, is challenging. To address these issues, we propose BALR-SAM, a boundary-aware low-rank adaptation framework that enhances SAM for medical imaging. It combines three tailored components: (1) a Complementary Detail Enhancement Network (CDEN) using depth-wise separable convolutions and multi-scale fusion to capture boundary-sensitive features essential for accurate segmentation; (2) low-rank adapters integrated into SAM's Vision Transformer blocks to optimize feature representation and attention for medical contexts, while simultaneously significantly reducing the parameter space; and (3) a low-rank tensor attention mechanism in the mask decoder, cutting memory usage by 75\% and boosting inference speed. Experiments on standard medical segmentation datasets show that BALR-SAM, without requiring prompts, outperforms several state-of-the-art (SOTA) methods, including fully fine-tuned MedSAM, while updating just 1.8\% (11.7M) of its parameters. The code is publicly available at \href{https://github.com/Misterursw/cad}{https://github.com/Misterursw/cad}.
\end{abstract}




\end{frontmatter}



\section{Introduction}
\label{sec1}
Medical image segmentation plays a crucial role in clinical workflows, which have attracted widespread attention from both industry and academia. However, achieving reliable segmentation in clinical environments is challenging due to several factors: ambiguous tissue boundaries, complex anatomical structures, and the limited availability of annotated data, which is costly and requires specialized expertise for labeling. Meanwhile, resource constraints in healthcare settings, such as hospitals with limited computational infrastructure, pose a distinct yet equally significant challenge~\cite{shen,ruan2025learning}. As a result, models that prioritize high diagnostic reliability, minimal parameters, fast inference, and low memory consumption are in high demand. To address this problem, \cite{melo} argue that adapting a generalized, large-scale pretrained model with minimal fine-tuning for diverse medical tasks provides greater practical value than training multiple task-specific models. Xiang et al.~\cite{xiang2025learning} propose a unified perspective called MMET for more robust visual-semantic embedding learning on generalizable image retrieval, which can further boost the performance of generalizable person re-identification. Although adopting a base model with domain-specific adaptations can align well with clinical needs for both efficiency and scalability, these approaches always fail to achieve satisfactory performance in certain specific fields. 

Recently, the Segment Anything Model (SAM)~\cite{sam}, pretrained on vast datasets of natural images, offers a promising foundation for such adaptation. Its ability to generalize across different segmentation tasks, supported by extensive pretraining, could help mitigate the scarcity of annotated medical data while leveraging its powerful feature extraction capabilities. However, SAM's performance in the context of medical imaging remains suboptimal without domain-specific tuning. Designed primarily for natural objects, SAM struggles with the subtle boundaries and complex textures inherent in anatomical structures, often leading to imprecise delineations that undermine its clinical utility. Moreover, its high computational overhead—manifested in significant memory requirements and slower inference speeds—clashes with the resource-efficient demands of real-world healthcare environment.

Recent efforts to adapt SAM for biomedical imaging highlight the growing interest in enhancing its utility. Various fine-tuning strategies have emerged, including parameter-efficient fine-tuning (PEFT) techniques like LoRA~\cite{lora}, which reduce the number of trainable parameters for task customization. Domain-specific adaptations such as $\mu$SAM~\cite{archit2023segment} and MedicoSAM~\cite{archit2025medicosam} improve baseline performance by pretraining SAM on microscopy and medical imaging data. Prompt-dependent approaches, including DeSAM~\cite{zhou2024cellseg1} and AI-SAM~\cite{10}, enhance segmentation with manual inputs like points or boxes. However, these advancements come with trade-offs: PEFT methods often compromise accuracy, prompt-based approaches limit automation, and fully fine-tuned models require excessive resources.

To address these limitations, we creatively propose a novel architecture named “BALR-SAM”, a comprehensive framework that synergistically integrates three novel components to enhance SAM's capabilities for medical image segmentation while maintaining computational efficiency. First, we introduce a Complementary Detail Enhancement Network (CDEN) that leverages depth-wise separable convolutions with dual-branch processing to capture fine-grained anatomical boundaries at multiple scales—a critical factor in medical imaging. Second, we implement strategically designed and positioned low-rank decomposition adapters within SAM's Vision Transformer blocks, reducing the parameter space by 94\% compared to full-rank alternatives while enabling precise domain adaptation. Third, we develop a low-rank tensor attention mechanism that fundamentally transforms the mask decoder's attention mechanism, reducing memory usage by approximately 75\% during training compared to SAM's original multi-head self-attention. Our extensive experiments on the standard medical segmentation dataset demonstrate that BALR-SAM could achieve state-of-the-art performance while updating just 1.8\% (11.7M) of its parameters.

The remainder of this paper is structured as follows. In Section~\ref{sec2}, we give the related works based on hand-crafted based approaches and deep learning based methods in medical area, and then briefly introduce our method. In Section~\ref{sec3}, the details of our Complementary Detail Enhancement Network, Low-Rank Decomposition Adapters, as well as the Low-Rank Tensor Attention Mechanism, are presented. Extensive evaluations compared with state-of-the-art methods and comprehensive analyses of the proposed approach are reported in Section~\ref{sec4}. Finally, conclusion of this paper and discussion of future works are presented in Section~\ref{sec5}.

\section{Related Work}
\label{sec2}
In this section, we give a brief review of the related works on common medical image segmentation methods. The core idea
of these existing methods is to efficiently model contextual information. These methods can be roughly divided into CNN-based approaches, Transformer-based approaches and SSM-based approaches.

\subsection{CNN-based Approaches}
Medical image segmentation methodologies have evolved through several distinct paradigms. Initially, Convolutional Neural Networks (CNNs) revolutionized the field. The Fully Convolutional Network~\cite{long2015fully} first introduced end-to-end architectures with skip connections. Subsequently, the U-Net~\cite{ronneberger2015unet} established the dominant encoder-decoder structure, which effectively captured multi-scale context while preserving localization detail, achieving strong performance even with limited data. This success spurred numerous variants. For instance, UNet++~\cite{zhou2018unetpp} utilized nested and dense skip connections to bridge the semantic gap between the encoder and decoder. Concurrently, Attention U-Net~\cite{oktay2018attention} integrated attention gates to refine feature maps by suppressing irrelevant regions. Moreover, 3D U-Net~\cite{cicek20163d} directly extended the architecture for volumetric data analysis. Beyond U-Net, other foundational CNN designs were adopted; ResNet~\cite{he2016deep} enabled deeper networks through residual learning, while DeepLabv3+~\cite{chen2018encoder} employed atrous convolutions for multi-scale object segmentation. Additionally, attention modules like CBAM~\cite{woo2018cbam} were incorporated to enhance feature representation. Nevertheless, CNNs possess an inherent limitation in modeling long-range dependencies due to their localized receptive fields, which motivated the exploration of alternative architectures.

\subsection{Transformer-based Approaches}
Transformers emerged as a powerful alternative, adept at capturing global contextual relationships. The Vision Transformer (ViT)~\cite{dosovitskiy2021image} first demonstrated applying a pure transformer to sequences of image patches. Following this, the Swin Transformer~\cite{liu2021swin} introduced a hierarchical structure and an efficient shifted-window self-attention mechanism. These architectures were quickly adapted for medical imaging. TransUNet~\cite{chen2021transunet}, for example, combined a transformer encoder with a CNN decoder to leverage both global context and precise localization. 
Guo et al.~\cite{guo2024uctnet} propose UCTNet where transformers focus on establishing global dependency for CNN’s unreliable regions.
Similarly, UNETR~\cite{hatamizadeh2022unetr} utilized a pure transformer encoder for 3D volumetric segmentation. Further refinements led to models like Swin UNETR~\cite{hatamizadeh2022swin}, which leveraged the Swin backbone, often with self-supervised pre-training, for state-of-the-art 3D segmentation. More recently, UNETR++~\cite{shaker2024unetrpp} proposed more efficient attention blocks to reduce the high computational cost associated with transformers. However, the quadratic complexity and substantial data requirements of transformers, especially for 3D volumes, remained significant challenges.

\subsection{SSM-based Approaches}

To address these efficiency concerns, State Space Models (SSMs) have recently gained prominence. SSMs offer a compelling approach to model long-range dependencies with linear computational complexity. The S4 model~\cite{gu2022efficiently} provided the theoretical foundation, which was later enhanced by Mamba~\cite{gu2023mamba} through a selective scan mechanism that enables content-based reasoning. These principles were translated to vision tasks in models such as Vision Mamba (Vim)~\cite{zhu2024vision} and VMamba~\cite{liu2024vmamba}, which introduced 2D scanning mechanisms. In medical segmentation, U-Mamba~\cite{ma2024umamba} integrated CNNs for local feature extraction with SSMs for global context modeling. In contrast, VM-UNet~\cite{ruan2024vmunet} proposed a pure SSM-based U-shaped architecture. Likewise, Mamba-UNet~\cite{wang2024mambaunet} synergized the U-Net framework with Mamba's efficient sequence modeling. SSMs are particularly advantageous for 3D data; for instance, SegMamba~\cite{xing2024segmamba} was specifically designed with a tri-oriented Mamba module for 3D volumetric segmentation. Consequently, SSMs present a highly efficient alternative to transformers, though they are a nascent technology with architectural best practices still rapidly evolving.

\section{Our Method}
\label{sec3}

\begin{figure}[!t]
\centering
\includegraphics[width=1.00\columnwidth]{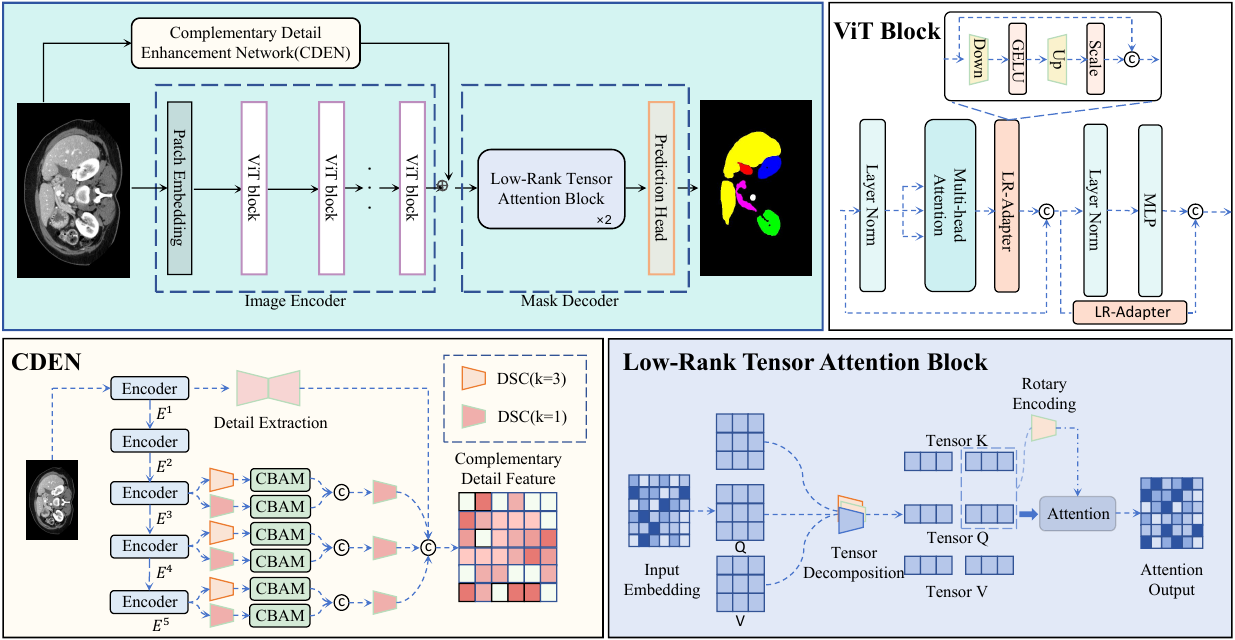}
\caption{Overview of the BALR-SAM. The image passing through the main branch, utilizing the Segment Anything Model’s (SAM) image encoder for initial feature extraction. Simultaneously, a complementary branch employs the Complementary Detail Enhancement Network (CDEN) to generate boundary-sensitive features, which are then fused with the image encoder’s features via element-wise addition. Within the image encoder’s Vision Transformer (ViT) blocks, specially designed low-rank adapters are strategically inserted at key positions—post multi-head self-attention and within the MLP residual path—to refine feature processing efficiently. The combined features subsequently feed into the mask decoder, where we overhaul the attention mechanism using a low-rank tensor decomposition of the query, key, and value matrices. Finally, the mask decoder’s prediction head outputs the predicted mask.}
\label{fig1}
\end{figure}

In this section, we first introduce the Complementary Detail Enhancement Network (CDEN) for generating boundary-sensitive features, followed by the Low-Rank Decomposition Adapters applied to the image encoder. Finally, we describe the Low-Rank Tensor Attention mechanism used in the mask decoder. The framework of proposed BALR-SAM is shown in Figure~\ref{fig1}. 

\subsection{Complementary Detail Enhancement Network}
Accurate boundary delineation in medical image segmentation is critical for identifying subtle anatomical structures, such as tumor margins or organ edges, which directly influence diagnostic precision. However, the Segment Anything Model (SAM) encoder, pretrained on natural images, often fails to capture these fine details in medical scans. To address this, we introduce the Complementary Detail Enhancement Network (CDEN), a lightweight module that enhances SAM’s encoder with boundary-sensitive features. CDEN leverages an efficient, multi-branch architecture to balance segmentation accuracy with the computational constraints typical of clinical environments. To be more specific,
CDEN processes input images through five encoding stages, progressively reducing resolution to extract multi-scale features. The first stage (E\textsuperscript{1}) preserves high-resolution spatial details—essential for outlining ambiguous boundaries, such as those in dermoscopic lesions or abdominal CT scans. To refine these features, we employ depth-wise separable convolutions (DSC)~\cite{dsc}, which reduce computational complexity by separating spatial and channel operations, making CDEN viable for resource-limited settings like small hospitals. Specifically, E\textsuperscript{1}’s output is processed with two DSC layers to emphasize fine edges often lost in deeper stages.

For the deeper stages (E\textsuperscript{3} to E\textsuperscript{5}), where features become more abstract, we adopt a dual-branch strategy to capture both local and global boundary information. One branch applies 1×1 convolutions to enhance local details (e.g., organ textures), while the other uses 3×3 convolutions to encode broader contexts (e.g., organ shapes). These complementary outputs are fused, enriching the encoder’s representation of complex anatomical structures with minimal overhead.

To prioritize clinically relevant features—such as lesion boundaries over background noise—we integrate the Convolutional Block Attention Module~\cite{cbam}. CBAM employs lightweight channel and spatial attention to focus on key regions, improving segmentation without significantly increasing resource demands. Finally, CDEN fuses features across all stages by upsampling deeper outputs (E\textsuperscript{3} to E\textsuperscript{5}) to align with E\textsuperscript{1}’s resolution, followed by a 1×1 convolution to adjust channels. This multi-scale integration ensures both detailed boundaries and contextual awareness, critical for tasks like multi-organ segmentation.

CDEN enhances SAM’s encoder by delivering precise boundary detection with reduced computational cost. In clinical applications, such as delineating skin cancer margins or segmenting abdominal organs, this efficiency translates to faster, more reliable results on standard hardware, supporting timely diagnosis in resource-constrained settings.

\subsection{Low-Rank Decomposition Adapters}
Beyond enhancing detail extraction as discussed in the previous section, we aim to better leverage the pre-trained information in the Segment Anything Model (SAM) image encoder for medical image segmentation. To achieve this, we introduce a novel low-rank adaptation strategy by strategically inserting adapters at optimal locations within the Vision Transformer (ViT) blocks and designing them with a specialized structure. This approach refines the image encoder’s attention mechanism to prioritize medically relevant features and enhances feature transformation in the MLP layers, all while preserving computational efficiency.

In each ViT block of the SAM image encoder, we integrate two low-rank adapters. The first is placed after the multi-head self-attention (MHSA) layer and before its residual connection, refining the attention mechanism to capture medical image-specific features. The second is positioned in the residual path of the multi-layer perceptron (MLP) layer following the MHSA, adjusting the feed-forward processing for domain-specific enhancement. These insertions maximize the reuse of the pre-trained encoder’s knowledge while introducing minimal overhead.

This placement targets medical imaging needs: MHSA captures global patterns but misses subtle boundaries (e.g., organ edges in CT scans), so the post-MHSA adapter refines attention for these details. The MLP enhances features like tumors or contrasts, often weak in SAM’s natural-image training, and its adapter boosts these signals, aligning the encoder with clinical demands efficiently.

On the contrary, traditional adapters rely on full-rank transformation matrices, $W \in \mathbb{R}^{m \times n}$, yielding a parameter count of $m \cdot n$. For a standard ViT with $m = n = 768$, this results in 589,824 parameters per layer, which is computationally expensive. We address this by reformulating the weight matrix via low-rank decomposition:

\begin{equation}
W \approx U \cdot V^T
\end{equation}

\noindent where $U \in \mathbb{R}^{m \times r}$ and $V \in \mathbb{R}^{n \times r}$, with $r \ll \min(m, n)$. For $r = 16$, the parameter count reduces to $(m + n) \cdot r = 24,576$, a 94\% decrease compared to full-rank adapters, preserving efficiency without sacrificing adaptability.

Each low-rank adapter comprises a down-projection layer to compress the input into a lower-dimensional space (rank $r$), a GELU activation for non-linearity, an up-projection layer to restore the original dimension, and a scaling factor to modulate the output. For an input $x \in \mathbb{R}^{m}$, the adapter output is:

\begin{equation}
\text{Adapter}(x) = s \cdot (U_{\text{up}} \cdot V_{\text{up}}^T) \cdot \text{GELU}((U_{\text{down}} \cdot V_{\text{down}}^T) \cdot x)
\end{equation}

\noindent where $U_{\text{down}}, V_{\text{down}}$ and $U_{\text{up}}, V_{\text{up}}$ are the low-rank factors for the down- and up-projections, and $s$ is a learnable scaling parameter. This design enables precise tuning of the pre-trained encoder, effectively adapting its attention and feature processing to medical images while keeping the parameter space compact.

\subsection{Low-Rank Tensor Attention Mechanism}

The multi-head self-attention (MHSA) mechanism in the Segment Anything Model (SAM) mask decoder incurs a computational complexity of $O(n^2d)$, where $n$ is the sequence length and $d$ is the embedding dimension, rendering it inefficient for high-resolution medical images. We propose an elegant and robust low-rank tensor attention mechanism to replace the conventional MHSA. By employing tensor decomposition and targeted positional encoding, our approach reduces complexity to $O(n(R_Q + R_K + R_V)d)$. This significantly reduces memory usage and moderately accelerates inference time, while preserving the decoder's ability to capture intricate anatomical dependencies.
This low-rank approach suits medical imaging by efficiently handling high-resolution scans (e.g., Synapse CTs), where SAM’s MHSA struggles with memory and speed. It preserves fine anatomical details—like organ overlaps or lesion margins—crucial for diagnosis, making it practical for clinical use.

Our method first computes the query ($Q$), key ($K$), and value ($V$) matrices from the input $x_t \in \mathbb{R}^{n \times d}$ using SAM’s pre-trained weights: $Q = x_t W_Q$, $K = x_t W_K$, and $V = x_t W_V$, where $W_Q, W_K, W_V \in \mathbb{R}^{d \times d}$. Instead of updating these weights, we treat $Q$, $K$, and $V$ as intermediate matrices and decompose them immediately after generation. The tensor-based factorization is defined as:

\begin{equation}
M_t^{(m)} = \frac{1}{R_m} \sum_{r=1}^{R_m} a_m^r(x_t) \otimes b_m^r(x_t), \quad m \in \{Q, K, V\}
\end{equation}
Here, $M_t^{(m)}$ represents the decomposed form of $Q$, $K$, or $V$ at position $t$, with $a_m^r(x_t) \in \mathbb{R}^{n \times r}$ and $b_m^r(x_t) \in \mathbb{R}^{d \times r}$ as learnable low-rank factors, and $R_m$ (e.g., $R_Q$, $R_K$, $R_V$) is the rank. After testing and balancing performance and efficiency, we chose a rank value of 8 to maintain an optimal trade-off between segmentation accuracy and computational complexity. Only these factors are updated during training, preserving the original weights. This reduces the complexity from $O(n^2d)$ in SAM’s MHSA to $O(n(R_Q + R_K + R_V)d)$, offering substantial savings for large $n$, as in high-resolution medical scans.

We further integrate Rotary Positional Embedding (RoPE) into the decomposed factors, applying it selectively to enhance spatial awareness:

\begin{equation}
\tilde{M}_t^{(m)} = \text{RoPE}_t\left(\frac{1}{R_m} \sum_{r=1}^{R_m} a_m^r(x_t) \otimes b_m^r(x_t)\right)
\end{equation}

RoPE is specifically applied to the key factors ($K_b = b_K^r$) and query factors ($Q_b = b_Q^r$), as these components are responsible for computing spatial relationships in the attention mechanism. This targeted application ensures the precise encoding of anatomical spatial relationships without introducing any additional overhead.

\section{Experiments}
\label{sec4}

\subsection{Dataset and Implementation Details}
\label{sec4.1}
In this study, we utilized the Synapse~\cite{synapse} dataset for multi-organ segmentation and ISIC17~\cite{isic17} dataset for skin lesion segmentation, and employed CVC-300~\cite{vazquez2017benchmark}, CVC-ColonDB~\cite{tajbakhsh2015automated} and ETIS~\cite{silva2014toward} datasets for endoluminal scene segmentation. To be more specific, the Synapse dataset consists of 30 abdominal CT cases with a total of 3,779 axial abdominal CT images, including annotations for 8 abdominal organs such as the aorta, kidneys, liver, and spleen. The ISIC17 dataset comprises 2,594 dermoscopic images with corresponding lesion segmentation masks. 
CVC-300~\cite{vazquez2017benchmark} consists of 60 polyp images
and the resolution of the images is 574 × 500, CVC-ColonDB~\cite{tajbakhsh2015automated} dataset contains 300 images with associated polyp masks obtained from 13 polyp video sequences
acquired from 13 patients, and ETIS~\cite{silva2014toward} dataset consists of 196 polyp images
and the resolution of the images is 1225 × 966.
These datasets were partitioned into training, validation, and test sets in a 7:1:2 ratio, ensuring a well-distributed evaluation of the model’s performance. The images were resized to 512$\times$512 pixels to standardize the input size for model training.

For the model training, we set the initial learning rate to 1e-4 and applied Cosine Annealing for learning rate decay throughout the training process. This approach allows for a smooth reduction in the learning rate, helping the model converge effectively. The model was trained for 40 epochs and the Dice Similarity Coefficient (DSC) was used as the evaluation metric for multi-organ and skin lesion segmentation performance. On the other hand, we employ five widely-used metrics in the field of polyp segmentation, i.e., Recall, Precision, Accuracy, Dice and mean
IoU (mIoU) to evaluate the model performances.
During training, we recorded the number of trainable parameters to assess the model’s complexity. All experiments were conducted on a single NVIDIA RTX A800 GPU, which provided sufficient computational power for efficient training.

\begin{table}[!t]
\centering
\caption{Performance comparison with state-of-the-art methods on ISIC17 and Synapse datasets for multi-organ and skin lesion segmentation task respectively. \textbf{Bold} indicates the best.}
\footnotesize
\setlength{\tabcolsep}{0.01mm}{
\begin{tabular}{l c c cccccccc c} 
\toprule
\multirow{2}{*}{\textbf{Methods}} & \multirow{2}{*}{\textbf{Params}} & \textbf{ISIC17}  & \multicolumn{9}{c}{\textbf{Synapse} } \\
 &  & \textbf{DSC} & \textbf{Aorta} & \textbf{GB} & \textbf{KL} & \textbf{KR} & \textbf{Liver} & \textbf{PC} & \textbf{SP} & \textbf{SM} & \textbf{Avg}\\
\midrule
UNet \cite{1} & 34.53M & 83.07 & 84.00 & 56.70 & 72.41 & 62.64 & 86.98 & 48.73 & 81.48 & 67.96 & 70.11 \\
AttnUNet \cite{2}& 34.88M & 83.66 & 82.61 & 61.94 & 76.07 & 70.42 & 87.54 & 46.70 & 80.67 & 67.66 & 71.70 \\
PolypPVt \cite{3} & 25.11M & 85.56 & 82.34 & 66.14 & 81.21 & 73.78 & 94.37 & 59.34 & 88.05 & 79.40 & 78.08 \\
SwinUNet \cite{4} & 27.17M & 83.97 & 81.76 & 65.95 & 82.32 & 79.22 & 93.73 & 53.81 & 88.04 & 75.79 & 77.58\\
nnUNet \cite{6} & 16.00M & 90.80 & \textbf{93.01} & 71.77 & 85.57 & 88.18 & \textbf{97.23} & 83.01 & 91.86 & \textbf{85.26} & 86.99\\
VM-UNet \cite{7} & 27.43M & 89.03 & 86.40 & 69.41 & 86.16 & 82.76 & 94.17 & 58.80 & 89.51 & 81.40 & 81.08\\
EMCAD \cite{8} & 26.76M & 85.95 & 88.14 & 68.87 & 88.08 & 84.10 & 95.26 & 68.51 & 92.17 & 83.92 & 83.63\\
SAM \cite{sam} & 636.00M & 81.6 & 82.41 & 59.63 & 80.24 & 69.48 & 83.47 & 57.28 & 78.61 & 61.27 & 71.55 \\
AI-SAM \cite{10} & - & 86.6 & 88.89 & 74.53 & 86.56 & 85.01 & 96.30 & 72.84 & 90.32 & 79.24 & 84.21 \\
SAMed \cite{11} & 18.81M &84.42  & 87.77 & 69.11 & 80.45 & 79.95 & 94.80 & 72.17 & 88.72 & 82.06 & 81.88 \\
MedSAM \cite{12} & 636.00M & 85.80 & 84.52 & 67.73 & 83.50 & 81.35 & 89.23 & 64.72 & 87.13 & 72.14 & 78.79\\
\midrule
BALR-SAM (Ours) & 11.7M & \textbf{91.14} & 90.47 & \textbf{76.97} & \textbf{90.40} & \textbf{88.71} & 96.74 & \textbf{83.79} & \textbf{91.88} & 84.73 & \textbf{87.96}\\
\midrule
P-values & \multicolumn{10}{c}{$<$ 1e-2 (DSC)} \\
\bottomrule
\end{tabular}}
\label{tab:contrast}
\end{table}

\begin{table}[!t]
\centering
\caption{Performance comparison with state-of-the-art methods on CVC-300, CVC-ColonDB and ETIS datasets for polyp segmentation task. \textbf{Bold} indicates the best.}
\footnotesize
\setlength{\tabcolsep}{2.39mm}{
\begin{tabular}{l c c c c c c} 
\toprule
Dataset & Method & Recall & Precision & Acc & Dice & mIoU \\
\midrule
\multirow{7}{*}[1.5ex]{CVC-300} & UNet~\cite{1} & 70.60 & 73.52 & 96.58 & 65.15 & 75.52 \\
& ResUNet~\cite{diakogiannis2020resunet} & 55.62 & 63.68 & 95.98 & 53.68 & 68.60 \\
& PraNet~\cite{fan2020pranet} & 86.80 & 85.34 & 98.56 & 86.31 & 88.56 \\
& SANet~\cite{zhang2021sa} & 87.00 & 87.56 & 98.55 & 87.85 & 90.21 \\
& UM-Net~\cite{du2025net} & 88.70 & 88.91 & 98.75 & 88.81 & \textbf{91.25} \\
& BALR-SAM (Ours) & \textbf{91.16} & \textbf{93.2} & \textbf{99.42} & \textbf{91.71} & 85.18 \\
\midrule
\multirow{7}{*}[1.5ex]{CVC-ColonDB} & UNet~\cite{1} & 59.73 & 72.85 & 94.44 & 54.77 & 70.00 \\
& ResUNet~\cite{diakogiannis2020resunet} & 46.66 & 64.65 & 93.59 & 45.39 & 64.09 \\
& PraNet~\cite{fan2020pranet} & 79.91 & 82.84 & 96.69 & 74.28 & 82.34 \\
& SANet~\cite{zhang2021sa} & 79.32 & 85.23 & 96.09 & 74.56 & 81.23 \\
& UM-Net~\cite{du2025net} & \textbf{80.32} & \textbf{85.58} & 96.86 & 76.08 & \textbf{82.82} \\
& BALR-SAM (Ours) & 77.78 & 84.03 & \textbf{97.63} & \textbf{79.39} & 82.58 \\
\midrule
\multirow{7}{*}[1.5ex]{ETIS} & UNet~\cite{1} & 53.10 & 47.67 & 91.66 & 34.52 & 59.85 \\
& ResUNet~\cite{diakogiannis2020resunet} & 46.49 & 30.64 & 90.80 & 27.80 & 55.30 \\
& PraNet~\cite{fan2020pranet} & 75.61 & 69.78 & 96.62 & 62.98 & 77.12 \\
& SANet~\cite{zhang2021sa} & 74.35 & 74.85 & 95.26 & 63.45 & 77.84 \\
& UM-Net~\cite{du2025net} & 75.45 & 75.60 & 96.95 & 64.48 & 78.52 \\
& BALR-SAM (Ours) & \textbf{80.70} & \textbf{80.52} & \textbf{98.91} & \textbf{79.30} & \textbf{79.79} \\
\bottomrule
\end{tabular}}
\label{tab2}
\end{table}

\subsection{Comparison with State-of-the-art Methods}
\textbf{Multi-Organ and Skin Lesion Segmentation.}
In this section, we compare our proposed method with state-of-the-art medical image segmentation approaches, including both classical segmentation models and recent SAM-based automatic (prompt-free) methods. Table~\ref{tab:contrast} presents the comprehensive comparison results on both the ISIC17 and Synapse datasets. Our experiments reveal several important findings. First, our parameter-efficient adaptation approach significantly outperforms the fully fine-tuned SAM, despite using only 4.2\% (11.7M) of SAM's trainable parameters. More impressively, our method surpasses even the fully fine-tuned MedSAM, which was specifically designed for medical imaging applications. This superior performance is consistent across both datasets, with particularly notable improvements in boundary precision and complex anatomical structure delineation.

\textbf{Polyp Segmentation.}
To comprehensively evaluate the effectiveness of our model in polyp segmentation, we compare the proposed BALR-SAM method with several existing polyp segmentation models: UNet~\cite{1}, PraNet~\cite{fan2020pranet}, ResUNet~\cite{diakogiannis2020resunet},  SANet~\cite{zhang2021sa} and UM-Net~\cite{du2025net} on CVC-300~\cite{vazquez2017benchmark}, CVC-ColonDB~\cite{tajbakhsh2015automated} and ETIS~\cite{silva2014toward} datasets, qualitative analysis results are presented in Table~\ref{tab2}.  It can be easily observed that the proposed BALR-SAM model achieves the highest mDice across the three unseen datasets, demonstrating strong generalization performance. Notably, on the ETIS dataset,
which differs significantly from the training set, our BALR-SAM model attains an Accuracy of 98.91 and  Dice of 79.30, surpassing the second-ranked UM-Net~\cite{du2025net} model by \textbf{+1.96\%} and \textbf{+14.82\%}, respectively.

\begin{table}[!t]
  \centering
  \caption{Ablation Study of our BALR-SAM method on the ISIC17 Dataset for skin lesion segmentation task.}
  \footnotesize
  \setlength{\tabcolsep}{4.7mm}{
    \begin{tabular}{cccccc}
    \toprule
    \multicolumn{4}{c}{Different Components of BALR-SAM} & \multicolumn{2}{c}{Metrics} \\
    \cmidrule(lr){1-4} \cmidrule(lr){5-6}
    Baseline & LR-Tensor  & LR-Adapters & CDEN  & mIoU$\uparrow$  & mDSC$\uparrow$ \\
    \cmidrule(lr){1-4} \cmidrule(lr){5-6}
    \checkmark &   &   &   & 14.3 & 21.4 \\ 
    \checkmark & \checkmark  &   &  & 64.8 & 74.6 \\
    \checkmark & \checkmark & \checkmark  &  & 77.2 & 85.6 \\
    \checkmark & \checkmark & \checkmark & \checkmark & 84.1 & 91.1 \\
    \bottomrule
    \end{tabular}}%
  \label{table:ablation}%
\end{table}%

\begin{figure}[!t]
\includegraphics[width=\textwidth]{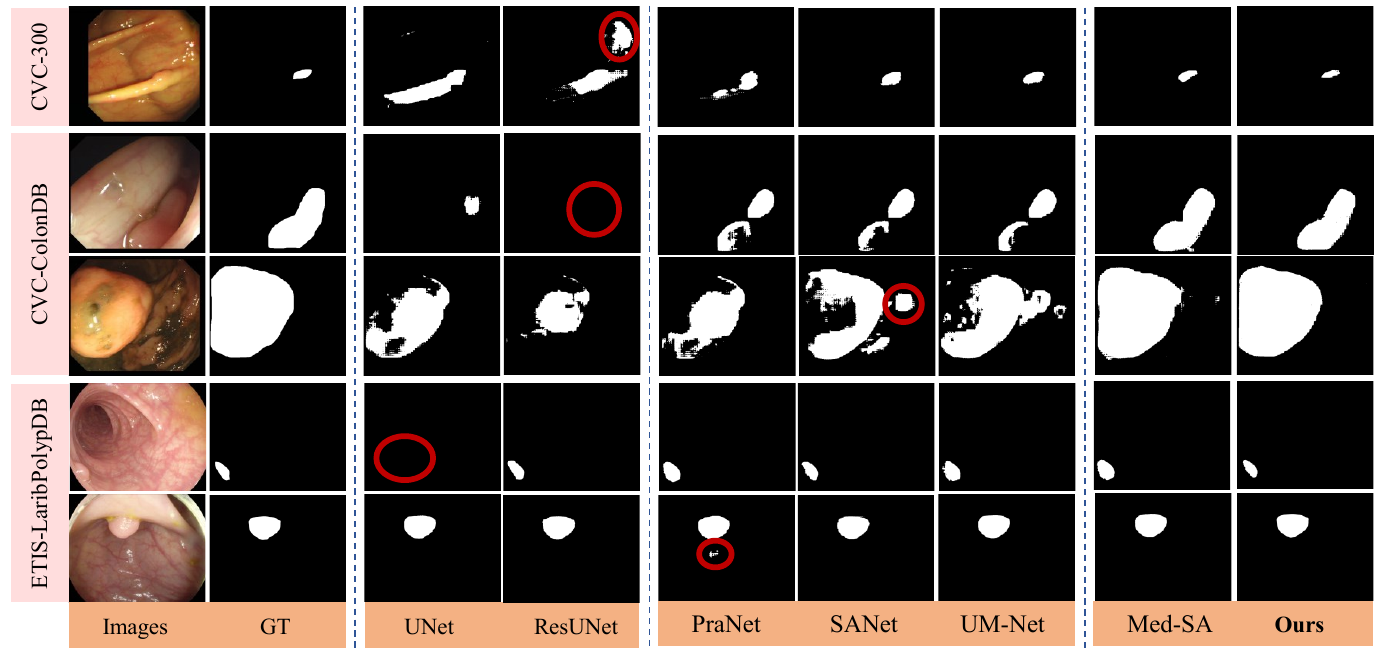}
\caption{The visual comparison results of our method on CVC300, ColonDB and ETIS datasets. GT represents ground truth.} 
\label{fig2}
\end{figure}

\subsection{Ablation Studies}
We conduct an ablation study to systematically evaluate the contribution of each component in our proposed method, the results are shown in Table \ref{table:ablation}. Starting with the SAM model as our baseline, we progressively incorporate our three key innovations on the ISIC17 dataset, measuring performance using mIoU and mDSC metrics. The results demonstrate that each component makes a meaningful contribution to the overall performance. The complete framework, with all three components working synergistically, achieves the highest segmentation performance, confirming the effectiveness of our approach in adapting SAM for medical image segmentation tasks. Additionally, when only replacing the original attention mechanism with LR-Tensor Attention and training this specific part, memory usage decreased from 43,747MB to 11,748MB, reflecting a reduction of approximately 75\%. In parallel, inference time was reduced from 196 seconds to 148 seconds. These results emphasize the superiority of the newly introduced attention mechanism.

\begin{figure}[!t]
\includegraphics[width=\textwidth]{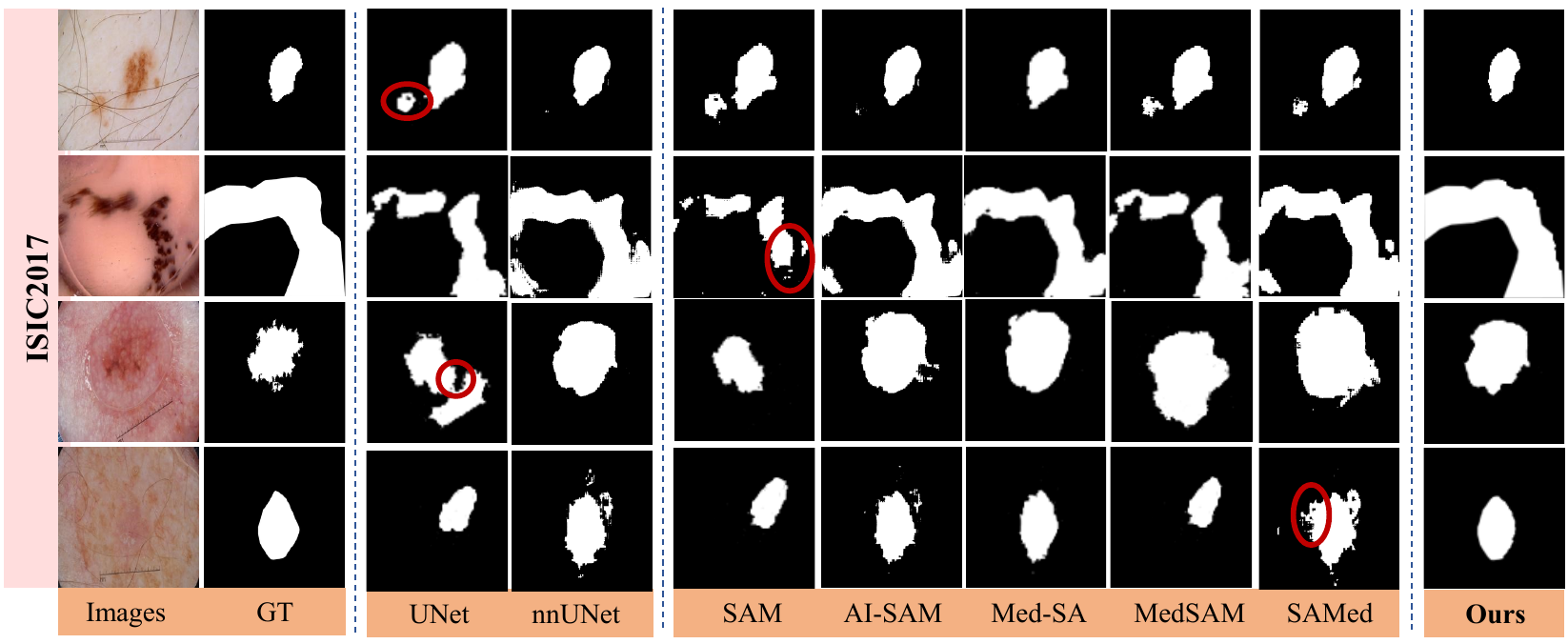}
\caption{The visual comparison results of our method on ISIC17 datasets. GT represents ground truth.} 
\label{fig3}
\end{figure}

\subsection{Visual results}
To visually demonstrate the effectiveness of our proposed method, we provide a visualization of the segmentation results for polyp segmentation in Figure~\ref{fig2}. To be more specific, the visual results of the traditional methods in Figure~\ref{fig2} reveals various segmentation errors, including redundancy, speckles, and omissions across different methods. On the contrary, the segmentation results of the proposed BALR-SAM method exhibit the highest overlap with the GT areas, closely resembling the ground truth. In the second and third rows of Figure~\ref{fig2}, the polyp tissues are relatively large and exhibit minimal differences from the surrounding healthy tissues, posing significant challenges for precise segmentation methods. Our BALR-SAM method utilizes Vision Transformer (ViT) as the network backbone, enabling a multi-level representation of image information, which results in better segmentation outcomes. For the small polyps in the first and fourth rows, our
method employs the  Complementary Detail Enhancement block to enhance the interaction between global and local information, allowing for more accurate identification and segmentation of small targets. The segmentation results in the fifth row demonstrate that our method can accurately segment the edges of polyps compared to other methods. This is attributed to the design of the SAM module in our proposed method, which selectively aggregates detail and semantic features, effectively improving the edge detection accuracy. In addition, similar improvements can also be easily observed in Figure~\ref{fig3} for skin lesion segmentation task.

\section{Conclusion}
\label{sec5}

This paper presents BALR-SAM, an efficient fine-tuning approach for adapting SAM to medical image segmentation. Our method combines a Complementary Detail Enhancement Network, low-rank adapters, and a low-rank tensor attention mechanism to surpass several SOTA methods while updating just 1.8\% of SAM's parameters. The clinical value of our approach lies in delivering high-quality segmentation with minimal computational demands, making advanced analysis accessible in resource-constrained healthcare settings. Future work will focus on better utilizing the prompt encoder to further improve performance and exploring the application of similar parameter-efficient adaptation techniques to vision-language models.



 \bibliographystyle{elsarticle-num} 
 \bibliography{pr-bibliography}



\end{document}